\RequirePackage{fix-cm}
\documentclass{svjour3}                     
\smartqed  
\usepackage{graphicx}
\usepackage{lineno,hyperref}
\usepackage{soul}
\soulregister\cite7
\soulregister\ref7
\soulregister\footnote7
\usepackage{color}
\usepackage{times}
\usepackage{epsfig}
\usepackage{graphicx}
\usepackage{amsmath}
\usepackage{amssymb}
\usepackage{algorithm}
\usepackage{algorithmic}
\usepackage{caption}
\usepackage{float}
\usepackage{flushend}
\usepackage{color}
\usepackage{subfig}
\usepackage{multirow}
\soulregister\cite7
\soulregister\ref7
\modulolinenumbers[5]
\usepackage{mathptmx}      
\journalname{Journal of Intelligent \& Robotic Systems}
\begin{document}
	
	\title{Stereo Frustums: A Siamese Pipeline for 3D Object Detection}
	
	\author{Xi Mo         \and
		Usman Sajid  \and
		Guanghui Wang
	}
	
	\institute{\textit{Frist Author}: Xi Mo \at
		Department of Electrical Engineering and Computer Science\\
		School of Engineering, University of Kansas \\
		Lawrence, Kansas, United States, 66045 \\
		\email{x618m566@ku.edu}
		\and
		\textit{Second Author}: Usman Sajid \at
		Department of Electrical Engineering and Computer Science\\
		School of Engineering, University of Kansas \\
		Lawrence, Kansas, United States, 66045 \\
		\email{usajid@ku.edu}
		\and
		\textit{Corresponding author}: Dr. Guanghui Wang \at
		Associate Professor \\
		Department of Computer Science\\
		Ryerson University\\
		Toronto, ON, Canada M5B 2K3\\
		\email{wangcs@ryerson.ca}
	}
	
	\date{Received: 7-19-2020 / Accepted: 10-27-2020}

	\maketitle
	
	\begin{abstract}
		The paper proposes a light-weighted stereo frustums matching module for 3D objection detection. The proposed framework takes advantage of a high-performance 2D detector and a point cloud segmentation network to regress 3D bounding boxes for autonomous driving vehicles. Instead of performing traditional stereo matching to compute disparities, the module directly takes the 2D proposals from both the left and the right views as input. Based on the epipolar constraints recovered from the well-calibrated stereo cameras, we propose four matching algorithms to search for the best match for each proposal between the stereo image pairs. Each matching pair proposes a segmentation of the scene which is then fed into a 3D bounding box regression network. Results of extensive experiments on KITTI dataset demonstrate that the proposed Siamese pipeline outperforms the state-of-the-art stereo-based 3D bounding box regression methods.
	\end{abstract}

	\section{Introduction} \label{section:intro}
	
	How to regress accurate 3D bounding boxes (bbox) for autonomous driving vehicles has become a pivotal topic recently. This technique can also benefit mobile robots and unmanned aerial vehicles with regard to scene understanding and reasoning. In this paper, we propose a Siamese pipeline method for 3D object detection.
	
	Given a pair of stereo images and the point cloud data collected by velodyne \cite{KITTI}, many approaches on a basis of deep-learning theories have been proposed to generate 3D bbox artifacts which can also be projected to a bird's-eye view (BEV) of LiDAR data for localization evaluation. According to the number of image views these approaches utilized, they can be divided into three categories: monocular view \cite{wang2019frustum,qi2018frustum,du2018general,shin2019roarnet,xu2018pointfusion,liang2018deep,ku2018joint,shi2019pointrcnn}, binocular views \cite{li2019stereo,Knigshof2019Realtime3D,wang2019pseudo,chen20173d,qin2019triangulation}, and non-view approaches \cite{Zhou2018,yang2018ipod,wang2019voxel,luo2018fast,yang2018pixor,yang2019std,shi2020points} that only processes point cloud. Mono-view based approaches focus on sensor-fusion of the camera and LiDAR sensors in either a global or a local manner, while non-view approaches extract point cloud features from hand-crafted voxels or raw coordinates. Compared to the extensive development in both categories mentioned above, there are fewer stereo-based and stereopsis-LiDAR-fusion works for 3D object detection. 
	
	Considering the runtime of stereo matching, coarse disparity map generated by fast stereo matching and GPU acceleration achieves real-time frame-rate, yet less accurate 3D detection results \cite{Knigshof2019Realtime3D} compared with that of coarse-to-fine disparity map \cite{wang2019pseudo}. However, it usually takes a few minutes to generate one panorama of coarse-to-fine disparity map before performing object detection tasks. Moreover, pixel-level stereo matching is sensitive to the error in the epipolar line calculated from camera calibration as stereo matching assumes all epipolar lines to be horizontal. We propose to reduce runtime by performing RoIs-level stereo matching instead of matching all pixels, and by a fast epipolar line searching strategy which calculates epipolar line from calibration data. We show in Section\ \ref{subsection:Runtime} that most of the runtime goes to point cloud processing.
	
\begin{figure}[t]
		\centering
		\includegraphics[width=0.45\linewidth]{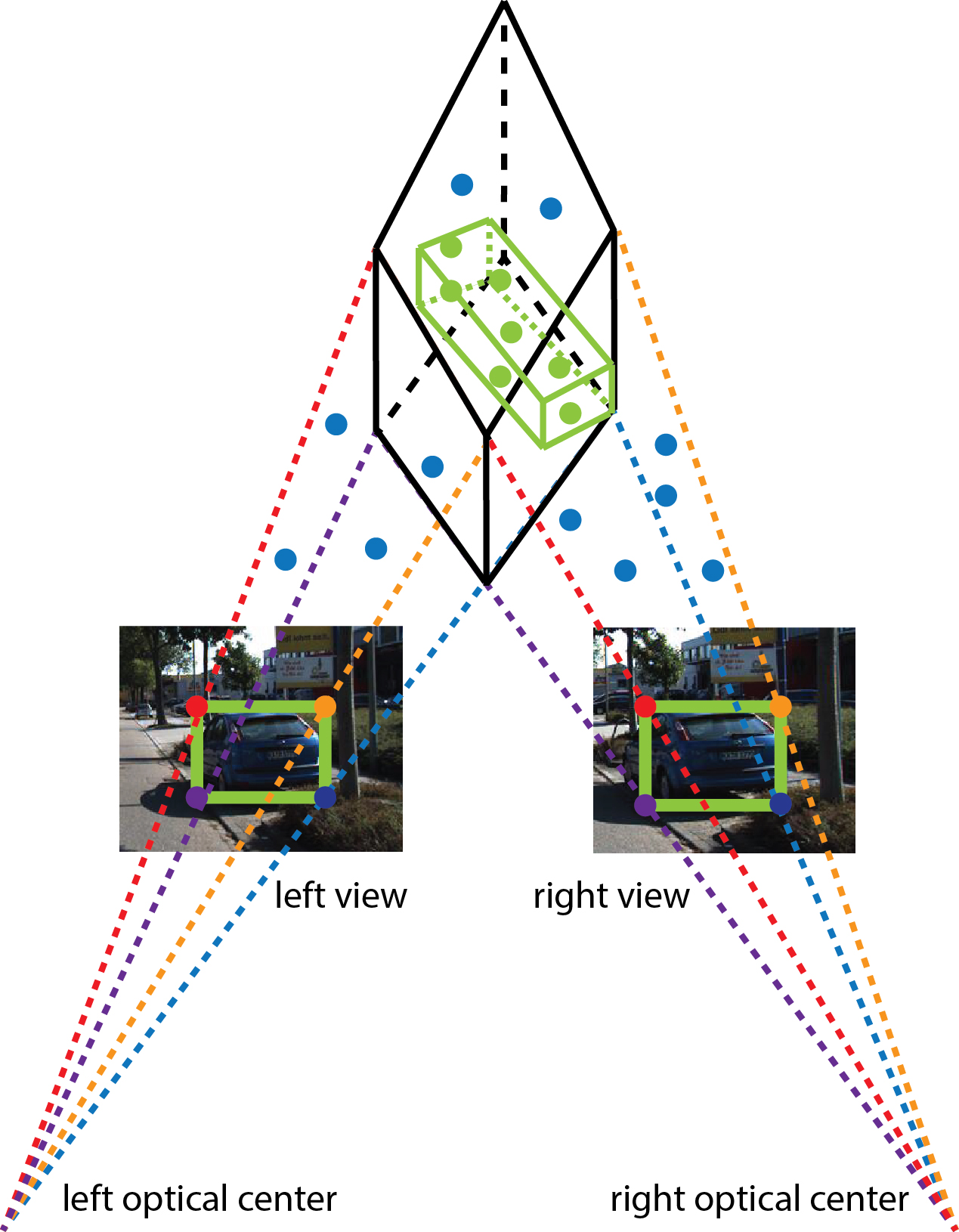}
		\DeclareGraphicsExtensions.
		\caption{Illustration of stereo frustums, valid inliers (green points) are segmented from inliers for predicting 3D bounding box (green). Given accurate 2D bboxes, the intersection of two frustums encloses the point cloud of the interested object with less ambiguity than single-frustum based and stereo-only methods, optimizing search space and being robust to perspective change.}
		\label{fig:1}
	\end{figure}

	Most stereo-based methods rely on stereo matching of stereopsis to generate depth maps for 3D object detection \cite{Knigshof2019Realtime3D,wang2019pseudo,chen20173d}, and one Stereo R-CNN based method \cite{li2019stereo} directly regresses keypoints of 3D bbox from the left-right correspondence of regional proposals (RoIs). In stereo matching, very close objects are usually located on the border area in both views with very large disparities. Therefore, part of the same object can be missing in either view. In this case, stereo-based methods may be unable to locate matched keypoints by stereo matching. Furthermore, considering the perspective changes, the same object may appear distinctively in both views due to occlusions. These situations may cause intrinsic ambiguities in stereo matching. As shown in Fig.\ \ref{fig:1}, the proposed method, by taking advantage of LiDAR-based 3D object detection methods, neither relies on stereo matching at pixel level nor predicts key points from corresponding RoIs.
	
	To circumvent the matching ambiguities, several approaches introduce spatial constraints as 3D anchors \cite{qin2019triangulation} and regress point cloud proposals transformed from dense disparity maps \cite{Knigshof2019Realtime3D,wang2019pseudo}. However, the design of a grid of anchors which has a proper density as well as high average precision (AP) for continuous space needs to be hand-crafted. Inspired by a novel single-frustum based method \cite{wang2019frustum}, we propose to solve the ambiguities by making full use of the spatial information for multi-modal regression. A novel module is proposed to directly map the point cloud onto the RoIs of the stereo image pairs and perform matching, by means of the normalized cross-correlation or the proposed 3D Intersection of Union (IoU) matching cost, and fast epipolar line search. This module correlates the 2D detection and synchronized 3D point cloud, and a novel network pipeline is proposed to accommodate this module. 
	
	The sparse nature of LiDAR points substantially reduces the number of points to be processed compared to the point cloud generated by the dense disparity map. Therefore, using LiDAR data can alleviate the computational burden of high-density matching at the pixel level. As to the perspective change, according to the setup of datum collecting vehicle, the baseline of stereo cameras is 0.54\textit m such that both views trap the same set of 3D keypoints from an object even though regions of the object appear distinctively. In addition, our method is robust to a slight disturbance on 2D bboxes (see Section\ \ref{subsection:Quantitative}).
	
	The main contributions of this paper include:
	\vspace{-0.5em}
	\begin{itemize}
		\setlength{\itemsep}{0pt}
		\setlength{\parsep}{0pt}
		\setlength{\parskip}{0pt}
		\item We propose an embedded light-weight matching module to generate 3D segmentation proposals by the RoIs from stereopsises.
		\item The proposed 3D IoU cost and epipolar line search algorithm are efficient in finding matches without Cython or GPU acceleration.
		\item The proposed Siamese architecture bridges the gap between stereopsis and real LiDAR points by integrating the point cloud segmentation network with 2D RoIs.
	\end{itemize}
	The proposed framework has been evaluated extensively on the KITTI dataset \cite{geiger2013vision}. The experimental results outperform previous stereo-based 3D bbox regression approaches. When testing on KITTI validation set, our method maintains as high AP on car detection as F-PointNets (FPN) \cite{qi2018frustum}, and outperforms FPN on pedestrian detection. The proposed approach runs on average at 2-3 frames per second.
	
	\section{Related Works}\label{section:related}
	\subsection{Monocular Pipeline for 3D Detection}\label{subsection:Monopipeline}
	
	Du \emph{et al.} \cite{du2018general} propose a general pipeline to detect cars from point cloud subsets constrained by monocular 2D detections. Three categories of normalized templates generalized from CAD models are fitted to 3D proposals in each subset. Each proposal is generated by RANSAC algorithm. The voxelized 3D proposals with the highest matching scores are fed to a two-stage refinement network. 
	
	Another improvement of the monocular pipeline is FPN for the purpose of regressing amodal 3D bbox including bbox sizes, orientation, and 3D bbox center. To our best knowledge, this work firstly proposes the concept of frustum which assigns bins and scores to point cloud subsets constrained by 2D detections. Instead of voxelizing the point cloud \cite{Zhou2018} before feeding it into a segmentation network followed by a T-Net to regress offset of 3D bbox center, the authors of \cite{qi2018frustum} design and apply the initial feature extraction networks PointNet(v1) \cite{qi2017pointnet} and PointNet++(v2) \cite{qi2017pointnet++} that learn point coordinates directly. In following sections, we denote FPN with PointNet++ backbone as FPNv2.
	
	RoarNet \cite{shin2019roarnet} points out that the performance of FPN degrades if the camera sensors and velodyne sensor are not synchronized. This work proposes a geometric agreement search by selecting the best projection from a 2D detection to its 3D bbox within single frustum with the help of spatial scattering to refine the location of 3D bbox. This improvement alleviates but can not solve the ambiguity of monocular back-projection (see Fig.\ \ref{fig:1}). Recently, F-ConvNet \cite{wang2019frustum} ranks leading position on KITTI benchmark. It proposes a sliding-window fashion along frustum to solve the localization ambiguity. As one of the state-of-the-art monocular detectors, F-ConvNet remedies improper hand-crafted divisions by concatenating point features from all windows, and learns valid objectness by a fully-convolutional network.
	
	\subsection{Stereo-Based Methods} \label{subsection:Stereobased}
	Several works have discussed the possibilities of 3D bbox regression by 2D detection w/o auxiliary depth information. Li \emph{et al.} \cite{li2019stereo} propose a new end-to-end approach based on stereo R-CNN to perform regional detection, and it is also integrated to a 2D-keypoint prediction network designed for vertex estimation of 3D bbox. Nevertheless, the predicted 2D-keypoints are lack of accuracy considering perspective changes. In addition, the orientation of the object is not included in the regression artifacts. 3DOP \cite{chen20173d} trains a structured SVM to generate 3D bbox proposals, which learns weights for an energy function that incorporates the point cloud density, free space, height prior, and height contrast information. However, without taking advantage of dense epipolar constraints in raw stereopsis, 3DOP predicts less accurate 3D bboxes than our approach. 
	
	Recently, a triangulation learning network \cite{qin2019triangulation} aims at learning epipolar constraints. The network requires anchors grid and 3D bbox ground truth to train. Left and right RoIs are selected by frustum-like forward-projection of 3D bboxes before cosine similarity is imposed on their feature maps. By cosine coherence scores computed from the left-right RoI pairs, the reweighting process weakens the signals from noisy channels. This method does not utilize dense raw epipolar constraints, and heavily relies on sparse anchor grids for localization. Also, this method brings in ambiguity since it searches among all potential anchors captured by a single frustum. 
	
	RT3D \cite{Knigshof2019Realtime3D} and Pseudo-LiDAR \cite{wang2019pseudo} estimate RGB-D image from stereopsis, then transform it into the point cloud, and regress 3D bbox with off-the-shelf clustering or LiDAR-based methods. RT3D presents a realtime detection scheme but comes with lower AP, while Pseudo-LiDAR predicts a less accurate depth map than real LiDAR data, which we will show in our experiments that, by the same FPN detector, our method achieves higher AP.    
	
	Another trend for object detection is through sensor fusion, i.e., to synchronize signals from multiple sensors. You \emph{et al.} \cite{you2019pseudo} proposed a scheme, namely PL++, to fuse sparse LiDAR points and the corresponding point cloud generated from the RGB-D image. As one of the top performers in 3D object detection, PL++ takes advantage of highly precise LiDAR points in localization. Zhang \emph{et al.} \cite{zhang2017joint} designed a deep network that fuses accurate depth maps, color images, and optical flow data. The model outperforms state-of-the-art works by a large margin. In order to further explore accurate RGB-D image in object detection, Tian \emph{et al.} \cite{tian2018robust} proposed a novel representation for a 2D convolutional network that encodes depth map, multi-order depth template, and height difference map. This approach achieves real-time performance, as well as being robust to insufficient illumination and partial occlusion.
	
	\begin{figure*}
		\centering
		\includegraphics[width=\linewidth]{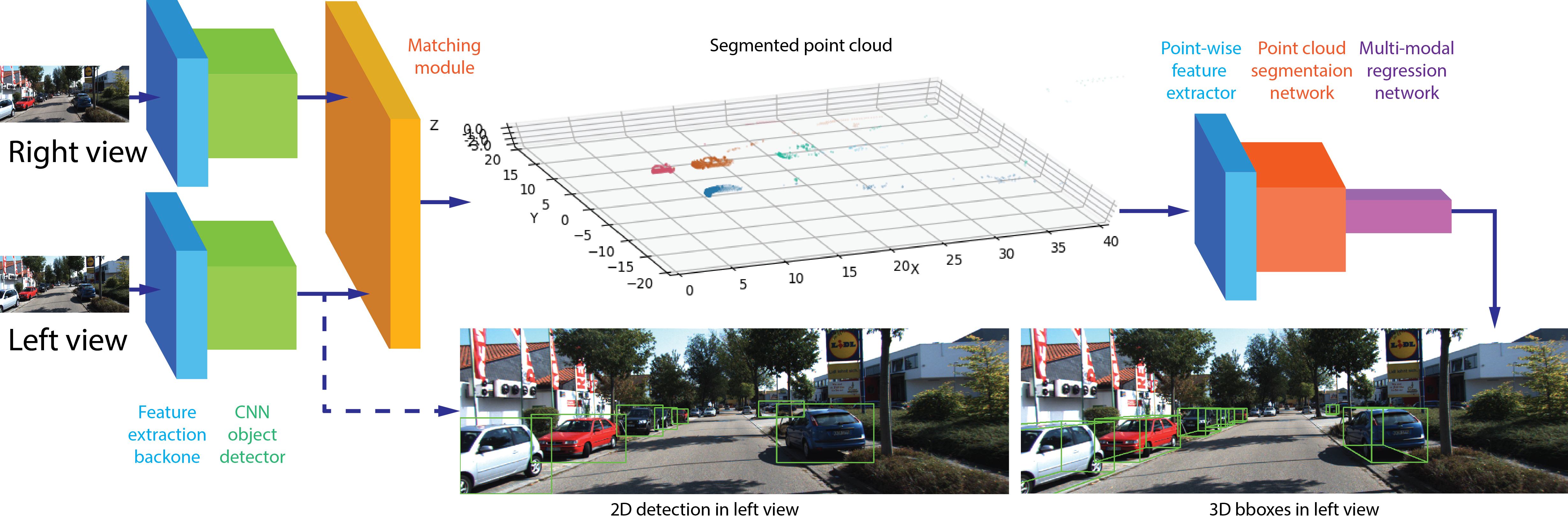}
		\DeclareGraphicsExtensions.
		\caption{Siamese pipeline for 3D object detection. Arrows signify flow directions, the dash line signifies an inspection of flow node. Feature extraction backbone and CNN object detector of sibling branches share the same weights and parameters.}
		\label{fig:3}
	\end{figure*}
	
	\section{Stereo Frustums Pipeline}\label{section:pipeline}
	
	In this section, we propose a Siamese pipeline (SFPN) that takes advantage of over-constrained epipolar geometry. Section \ref{subsection:Epconstraint} describes the dense epipolar constraints that SFPN bases on. Section \ref{subsection:Sfpipelne} introduces an overview of stereo frustum pipeline. Section \ref{subsection:Matchalg} introduces four RoIs matching methods that our module has implemented. In this section, we assume the 2D object detections on both views are consistent with their ground truths. 
	
	\begin{figure}[htbp]
		\centering
		\includegraphics[width=\linewidth]{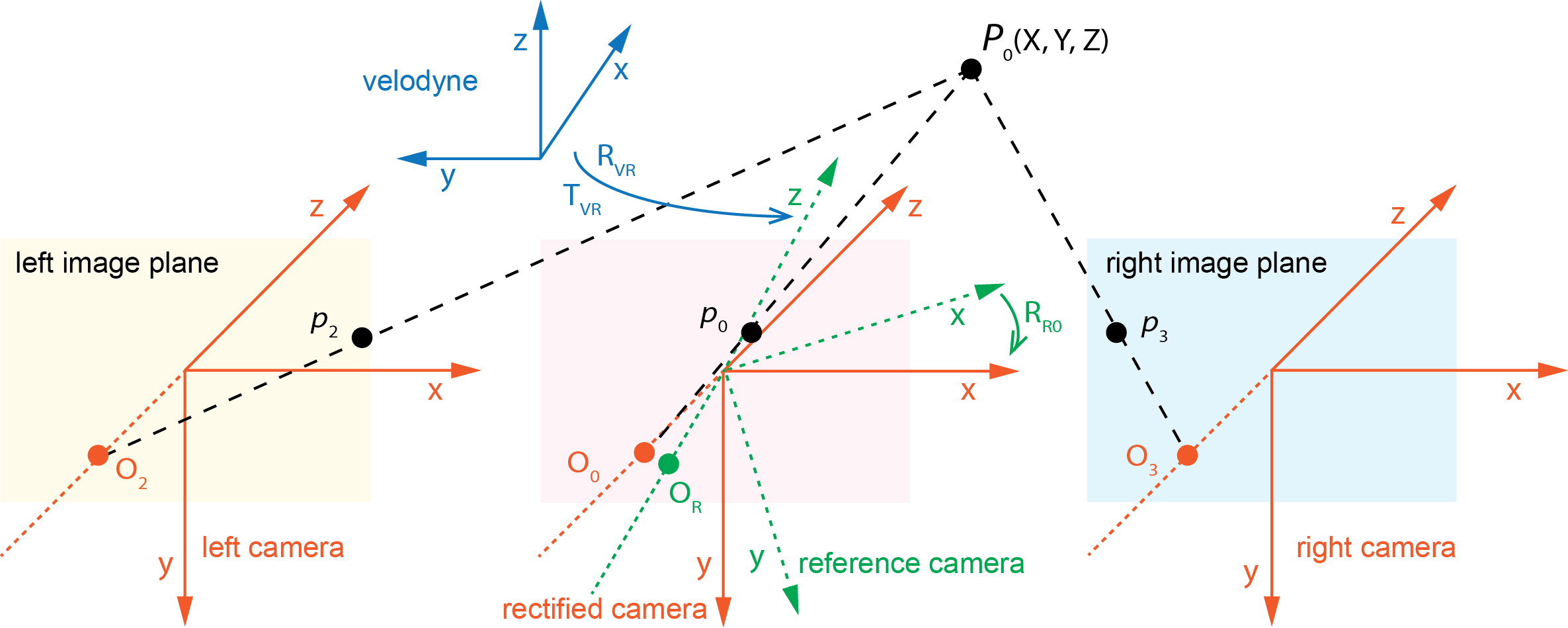}
		\DeclareGraphicsExtensions.
		\caption{Coordinate systems of the KITTI data-collection vehicle \cite{geiger2013vision}. $P_0$ is a LiDAR point in velodyne coordinate system (blue), its forward-projections in left view and right view are denoted as $p_2$ and $p_3$ respectively. $O_2$ and $O_3$ are the optical centers (orange) of left color camera and right color camera, $O_0$ (orange) is the optical center of rectified camera coordinate system (gray-scale camera 0), and $O_R$ (green) is the optical center of reference camera coordinate system. Rotation from reference camera system (green) to rectified camera system (orange) is denoted as $R_{R0}$.}
		\label{fig:7}
	\end{figure}
	
	\subsection{Dense Epipolar Constraints}\label{subsection:Epconstraint}
	\textbf{Notations}.\hspace{0.5cm}Let $F \in \mathbb{R}^{3 \times 3}$ be the fundamental matrix defined by the left-right camera coordinate systems (see Fig.\ {\ref{fig:7}}) whose optical centers are $O_2$ and $O_3$ respectively. Let $\mathcal{S}_2=\{l_i|i=1,2,...,m\}$ be the set of bboxes centered at $l_i$ in the left view, $\mathcal{S}_3=\{r_i|i=1,2,...,n\}$ be the set of bboxes centered at $r_i$ in the right view, $P_c$ denotes the point cloud set of scene, with $P_c(l_i)$ a subset of $P_c$ by forward projecting $P_c$ to the bbox region centered at $l_i$, $N_{thres}$ denotes the minimum number of spacial points required for the multi-modal regression network, and $N(P_c(l_i))$ denotes the number of inliers of $P_c(l_i)$. SPFN is valid if the following two constraints are satisfied:\vspace{0.5em}
	
	\noindent \textbf{(1) One-to-one onto mapping}. $\exists \mathcal{S}_2^{'}\subseteq \mathcal{S}_2$ and $\exists\mathcal{S}_3^{'}\subseteq \mathcal{S}_3$ subject to, $\forall l_i\in\mathcal{S}_2^{'},\ \exists r_j\in\mathcal{S}_3^{'}$ such that $r_j^TFl_i=0$, and $\forall r_j\in\mathcal{S}_3^{'},\ \exists l_i\in\mathcal{S}_2^{'}$ such that $r_j^TFl_i=0$.
	
	\noindent\textbf{(2) Minimum intersection}. Given conditions (1), suppose $N_{sec}$ is number of inliers in the stereo-frustums intersection as shown in Fig.\ {\ref{fig:1}}, $\forall l_i\in\mathcal{S}_2^{'},\ \exists r_j\in\mathcal{S}_3^{'}$ subject to $r_j^TFl_i=0$, then $N_{sec}=N(P_c(l_i)\cap P_c(r_j))\geq N_{thres}$.\vspace{0.5em}
	
	The one-to-one onto mapping constraint finds all spacial points that are forward-projected to matched the left and right RoIs, and the minimum intersection constraint is designed for the multi-modal regression network, ensuring that the regression will not fail due to too few spacial points as the input. It is widely known that the initial step after installing mounted devices is to get calibration information for all sensors including camera sensors, velodyne sensors, and IMU sensors. Then, fine-tune all sensors according to the calibration data until an acceptable error rate is observed. For the KITTI dataset, we assume the data collection system is well-calibrated, thus calibration files are reliable enough.
	
	A necessary constraint for stereo matching based approaches is that, the stereo search can be performed horizontally. To relax this constraint, it is important to estimate the epipolar constraints from calibration files. In Section \ref{subsection:Matchalg}, we employ a searching algorithm based on the estimated epipolar constraints other than the horizontal search strategy.\\
	
	\noindent\textbf{Fundamental matrix}.\hspace{0.2cm} In KITTI dataset, the fundamental matrix between the left and the right image planes $I_2,I_3$ can be calculated as:
	\begin{equation}\label{eqn:1}
	F=K_{c3}^{-T}[\textbf{t}_{c2}^{c3}]_\times K_{c2}^{-1},
	\end{equation}
	where $K_{c2}, K_{c3}$ are non-singular $3\times{3}$ intrinsic matrices of the left and right cameras respectively, $\textbf{t}_{c3}^{c2}$ is a 3D translation vector\footnote{Each pair of subscript and superscript indicates the from-to relation, the same notation applies to other translation vectors and rotation matrices in this paper.} from the left to the right camera coordinate system (see Fig.\ \ref{fig:7}), $[\mathbf{\cdot}]_\times$ indicates the antisymmetric matrix of a vector. Using homogeneous coordinates of image points, $\forall p_3\in{I_3}$ and its correspondent image point $p_2\in{I_2}$ subject to $p_3^{T}Fp_2=0$, by which the epipolar line through $p_3$ or $p_2$ can be deduced directly. To compute $\textbf{t}_{c2}^{c3}$ of Eq. (\ref{eqn:1}), according to the KITTI calibration methodology  \cite{geiger2013vision}, we know the projection from a 3D point $\tilde{P}_0$ to its image $\tilde{p}_0$ in each image plane can be denoted as $\tilde{p}_i=[K_{ci}|\textbf{C}_i]\tilde{P}_0=P_{rect}^{(i)}\tilde{P}_0, i\in\{0,1,2,3\}$, where $\textbf{C}_i$ is the last column of projection matrix $P_{rect}^{(i)}$, and $i$ denotes the $i$-th camera: value `2' stands for the left camera, `3' for the right camera, and `0' for the gray-scale camera selected as the rectified camera. Then, we have the result
	\begin{equation}\label{eqn:2}
	\textbf{t}_{c2}^{c3} = K_{c2}^{-1}\textbf{C}_2-K_{c3}^{-1}\textbf{C}_3.
	\end{equation}	 
	
\paragraph{Proof} The coordinate systems of KITTI data-collection vehicle are depicted in Fig.\ \ref{fig:7}. Assume $P_0$ is a inhomogeneous-coordinate point in rectified camera coordinate system, its images in left and right image planes are $p_2$ and $p_3$ (homogeneous-coordinates), respectively. According to the pinhole model, we have:
\begin{equation}\label{eqn:3}
p_3 = K_{c3}(R_{c0}^{c3}P_0+\textbf{t}_{c0}^{c3}) \Rightarrow p_3 = [K_{c3}R_{c0}^{c3}|K_{c3}\textbf{t}_{c0}^{c3}]P_0,
\end{equation}
where $R_{c0}^{c3}$ and $\textbf{t}_{c0}^{c3}$ are $3\times{3}$ rotation matrix and $3\times{1}$ translation vector from the rectified camera coordinate system to right camera coordinate system. Likewise, 
\begin{equation}\label{eqn:4}
p_2 = K_{c2}(R_{c0}^{c2}P_0+\textbf{t}_{c0}^{c2}) \Rightarrow p_2=[K_{c2}R_{c0}^{c2}|K_{c2}\textbf{t}_{c0}^{c2}]P_0.
\end{equation}
According to the projection matrix $P_{rect}^{(i)}$ mentioned above, we know $p_i=[K_{ci}|\textbf{C}_i]P_0$. Comparing the projection matrices $[K_{c2}|\textbf{C}_i]$ and $[K_{c3}|\textbf{C}_2]$ with the projection matrices in Eq. (\ref{eqn:3}) and Eq. (\ref{eqn:4}), respectively, we have the following observations:  
\begin{equation}\label{eqn:5}
R_{c0}^{c3}=R_{c0}^{c2}=I\ \Rightarrow\ R_{c2}^{c3}=I,\ \textbf{t}_{c0}^{ci}=K_{ci}^{-1}\textbf{C}_i, i=2,3,
\end{equation}
which indicates the rotation between the left camera and right camera can be ignored. It can be derived from Eq. (\ref{eqn:5}) that
\begin{equation}\label{eqn:6}
\textbf{t}_{c2}^{c3}=\textbf{t}_{c0}^{c2} - \textbf{t}_{c0}^{c3} = K_{c2}^{-1}\textbf{C}_2-K_{c3}^{-1}\textbf{C}_3,\ \mathrm{QED}.
\end{equation}	 

It is necessary to prove Equ. (\ref{eqn:1}) as well.  Consider Eq. (\ref{eqn:3}), Eq. (\ref{eqn:4}) again, we have
\begin{equation}\label{eqn:7}
K_{c3}^{-1}p_3=R_{c0}^{c3}P_0+\textbf{t}_{c0}^{c3},\ K_{c2}^{-1}p_2=R_{c0}^{c2}P_0+\textbf{t}_{c0}^{c2},
\end{equation}
and by using the observation $R_{c0}^{c3}=R_{c0}^{c2}=I$ from Eq. (\ref{eqn:5}) and eliminating $P_0$, we have 
\begin{equation}\label{eqn:8}
K_{c3}^{-1}p_3-\textbf{t}_{c0}^{c3}=K_{c2}^{-1}p_2-\textbf{t}_{c0}^{c2}\ \Rightarrow\ K_{c2}^{-1}p_2-K_{c3}^{-1}p_3=\textbf{t}_{c2}^{c3}.
\end{equation}
Premultiply $[\textbf{t}_{c2}^{c3}]_\times$ to both sides of Eq. (\ref{eqn:8}): 
\begin{equation}\label{eqn:9}
[\textbf{t}_{c2}^{c3}]_\times K_{c2}^{-1}p_2=[\textbf{t}_{c2}^{c3}]_\times K_{c3}^{-1}p_3,
\end{equation}
by premultiplying $(K_{c3}^{-1}p_3)^T$ to both sides of Eq. (\ref{eqn:9}), we have
\begin{equation}\label{eqn:10}
(K_{c3}^{-1}p_3)^T[\textbf{t}_{c2}^{c3}]_\times K_{c2}^{-1}p_2=0.
\end{equation}
which can be rearranged as
\begin{equation}\label{eqn:11}
p_3^T(K_{c3}^{-T}[\textbf{t}_{c2}^{c3}]_\times K_{c2}^{-1})p_2=0.
\end{equation}
It is obvious that $F=K_{c3}^{-T}[\textbf{t}_{c2}^{c3}]_\times K_{c2}^{-1}$, and $\textbf{t}_{c2}^{c3}$ can be calculated by Eq. (\ref{eqn:6}), QED.
	
	\subsection{Pipeline For Stereo Frustums} \label{subsection:Sfpipelne}
	
	Each matched RoIs pair from left-right views encodes dense epipolar constraints between pixels and their corresponding 3D spacial points. Though fewer LiDAR points are forward-projected to front-view, we observe that $\sim$10,000 points are captured by most bboxes. Among these points, there are many outliers that do not belong to the interested objects. As pointed out by Zhou \emph{et al.} \cite{Zhou2018} that it enforces high computational cost to FPN, our matching module effectively filters out 15\% to 49\% of points by dense epipolar constraints and thus can greatly speed up the networks that process point cloud.
	
	The proposed Siamese pipeline is shown in Fig.\ \ref{fig:3}. The stereopsis is fed to a 2D object detector. The detector can be any of traditional ones, such as the CNN-based detectors as Faster R-CNN \cite{ren2015faster}, RetinaNet \cite{lin2017focal}, the leading PC-CNN-V2 \cite{du2018general} on KITTI object detection benchmark, and others \cite{ma2020mdfn,li20202}. An accurate detector is crucial before populating the RoIs pairs into the matching module. Theoretically, each RoI in the left view should have its counterpart in the right view since one-to-one onto mapping constraint will be inconsistent otherwise. Also, If one salient object fails to be detected simultaneously in both views, it would lower both detection precision and recall. Note that only the objects in the left view have their ground truths, thus we focus on matching RoIs of the right view to the left RoIs. The matching module follows the one-to-one mapping constraint or its relaxed form and takes all right RoIs as candidates. The goal is to find all matches to objects in the left view by the matching costs, which is illustrated in Section\ \ref{subsection:Matchalg}.
	
	In the light of valid pairs proposed by the matching module and forward-projection matrices, the scene is segmented into 3D RoIs subject to minimum intersection constraint, whose semantic information and objectness scores are inherited from the 2D detections. A segmentation network refines the 3D RoIs with the extracted point-wise features. Furthermore, multi-modal regressor estimates the orientation, sizes of 3D bbox, and localization. To verify the validity of SFPN, we employ Faster R-CNN and Mask R-CNN \cite{he2017mask} as the 2D detectors, FPNv2 as the point cloud segmentation and multi-modal regression networks.
	
	\subsection{Matching Algorithms} \label{subsection:Matchalg}
	
	\textbf{Notations}.\hspace{0.5cm}Consider the centers of bboxes as potential matches, let $D_{thres}$ be the threshold of distance to the epipolar line, $P_{thres}$ the threshold of regional similarity, $P_{l_i}$ is the RoI image patch centered at $l_i$, $Prob_{ij}=NCC(P_{l_i},P_{r_j})$ denotes the normalized cross-correlation operator, $dist(r_j, e_i)$ is the operator to compute the distance from $r_j$ to the epipolar line $e_i$ of $l_i$, $IoU_{ij}=N(P_c(l_i)\cap P_c(r_j))/N(P_c(l_i)\cup P_c(r_j))$ as 3D IoU matching cost, $P_{3d\_thres}$ the probability threshold of the 3D IoU cost. We then formulate four matching algorithms as:\vspace{0.5em}
	
	\noindent\textbf{(1)} RoIs matching by 3D IoU cost and epipolar line search (3DCES, Alg.\ \ref{alg:3DCES}), which relaxes the one-to-one onto mapping constraint to left view only for alleviating computational burden.\vspace{0em}
	
	\noindent\textbf{(2)} RoIs matching by 3D IoU cost and method of exhaustion (3DCME, Alg.\ \ref{alg:3DCME}), which relaxes the one-to-one onto mapping constraint.\vspace{0em} 
	
	\noindent\textbf{(3)} RoIs matching by regional similarity cost (RSC, Alg.\ \ref{alg:RSC}), which relaxes the one-to-one onto mapping constraint.\vspace{0em}
	
	\noindent\textbf{(4)} RoIs matching by regional similarity cost and left-right consistency check (RSCCC,\ Alg.\ \ref{alg:RSCCC}), which strictly follows the one-to-one onto mapping constraint.\vspace{0.5em}
	
	The epipolar line constraint facilitates searching for a fast pixel-level matching. As for RoI-level matching, we adapt this approach for faster search to RSC, RSCCC, and 3DCES. Experiments on both regional similarity and 3D IoU costs show the effectiveness of epipolar line search. The matching module is designed using Python without any Cython and GPU acceleration, or paralleled programming. On account of the compatibility purpose, 3DCME and 3DCES merely return matches as RSC and RSCCC do. We will show in the experiments that SFPN directly processes raw point cloud with high efficiency, and RSC achieves the shortest runtime among the four modules. As previous works done by Li \emph{et al.} \cite{li2019stereo} and Qin \emph{et al.} \cite{qin2019triangulation} have designed learning-based methods to find RoIs matches as RSC and RSCCC, we are seeking the possibilities to design learning-based 3DCME and 3DCES.
	
	In order to deploy the most appropriate module for different tasks, it is recommended to inspect the groud-truth of the labeled dataset. In terms of a stereo camera configuration, if the groud-truths of both views are provided, then RSCCC may be implemented for the reason that 2D detection in both views can be described as `reliable', which reduces unreliable matches by left-right consistency check while maintains high efficiency in RoIs matching. If the groud-truth of one view is unavailable, the view without ground-truth is thereby unreliable. In this case, although implementation of RSC is the fastest in runtime, 3DCES may be the best choice as a trade-off of runtime and detection performance. Moreover, 3DCME is better implemented in case of a very sparse LiDAR scene for its ability to find accurate matches of frustums. Reader may refer to Section \ref{subsection:Quantitative} for comparisons on the performance of the proposed matching modules and Section \ref{subsection:Runtime} for details on the runtime analysis.
	
	\begin{algorithm}[htb] 
		\caption{3DCES} 
		\label{alg:3DCES} 
		\begin{algorithmic}[1] 
			\REQUIRE ~~\\ 
			$\mathcal{S}_2,\mathcal{S}_3,P_{thres},D_{thres}, P_c$;\\
			\ENSURE ~~\\ 
			\STATE Compute set of epipolar lines $\mathcal{E}=\{Fl_i|\forall l_i\in\mathcal{S}_2\}$; 
			\STATE $\forall l_i\in\mathcal{S}_2$, search leftwards in right view along $e_i\in\mathcal{E}$, calculate $\mathcal{R}_i=\{r_j|\forall r_j\in\mathcal{S}_3,\ dist(r_j, e_i)<D_{thres}\}$;
			\STATE Compute costs $\mathcal{C}_i=\{IoU_{ij}|\forall r_j\in\mathcal{R}_i, \exists IoU_{ij}\neq 0\}$;
			\STATE Find match. If $\mathcal{C}_i\neq\varnothing$ and $\exists r_j\in\mathcal{R}_i\ \mathrm{s.t.}\ IoU_{ij}\geq max(P_{3d\_thres},max(\mathcal{C}_i))$, let $m_i=(l_i, r_i)$;\\
			\RETURN $\mathcal{M}$; 
		\end{algorithmic}
	\end{algorithm}
	
	\begin{algorithm}[htb] 
		\caption{3DCME} 
		\label{alg:3DCME} 
		\begin{algorithmic}[1] 
			\REQUIRE ~~\\ 
			$\mathcal{S}_2,\mathcal{S}_3,P_{3d\_thres},P_c$;\\
			\ENSURE ~~\\ 
			\STATE $\forall l_i\in\mathcal{S}_2$, let $\mathcal{C}_i=\{IoU_{ij}|\forall r_j\in\mathcal{S}_3, \exists IoU_{ij}\neq 0\}$;
			\STATE Find match. If $\mathcal{C}_i\neq\varnothing$ and $\exists r_j\in\mathcal{S}_3\ \mathrm{s.t.}\ IoU_{ij}\geq max(P_{3d\_thres},max(\mathcal{C}_i))$, let $m_i=(l_i, r_i)$;\\
			\RETURN $\mathcal{M}$; 
		\end{algorithmic}
	\end{algorithm}
	
	\begin{algorithm}[htbp] 
		\caption{RSC} 
		\label{alg:RSC} 
		\begin{algorithmic}[1] 
			\REQUIRE ~~\\ 
			$\mathcal{S}_2,\mathcal{S}_3,P_{thres},D_{thres}$;\\
			\ENSURE ~~\\ 
			\STATE Compute set of epipolar lines $\mathcal{E}=\{Fl_i|\forall l_i\in\mathcal{S}_2\}$; 
			\vspace{0em}
			\STATE $\forall l_i\in\mathcal{S}_2$, search leftwards in right view along $e_i\in\mathcal{E}$, calculate $\mathcal{R}_i=\{r_j|\forall r_j\in\mathcal{S}_3,\ dist(r_j, e_i)<D_{thres}\}$;
			
			\STATE Perform RoI alignment to $P_{l_i}$ and $P_{r_j}$, compute set of costs $\mathcal{C}_i=\{Prob_{ij}|\forall r_j\in\mathcal{R}_i, \mathcal{R}_i\neq\varnothing\}$; 
			
			\STATE Find match. If $\mathcal{C}_{i}\neq\varnothing$ and $\exists r_j\in\mathcal{R}_i\ \mathrm{s.t.}\ Prob_{ij}\geq max(P_{thres},max(\mathcal{C}_i))$, let $m_i=(l_i, r_i)$;
			\RETURN $\mathcal{M}$; 
		\end{algorithmic}
	\end{algorithm}
	
	\begin{algorithm}[htbp] 
		\caption{RSCCC} 
		\label{alg:RSCCC} 
		\begin{algorithmic}[1] 
			\REQUIRE ~~\\ 
			$\mathcal{S}_2,\mathcal{S}_3,P_{thres},D_{thres}$;\\
			Matches of left-to-right RoIs pairs, $\mathcal{M}_{l}=\varnothing$;\\
			Matches of right-to-left RoIs pairs, $\mathcal{M}_{r}=\varnothing$;
			\ENSURE ~~\\ 
			\STATE Compute sets of epipolar lines\\$\mathcal{E}_r=\{Fl_i|\forall l_i\in\mathcal{S}_2\}$ and $\mathcal{E}_l=\{r_i^TF|\forall r_i\in\mathcal{S}_3\}$;\\
			\STATE $\forall l_i\in\mathcal{S}_2$, search leftwards in right view along $e_i\in\mathcal{E}_r$, calculate $\mathcal{R}_i=\{r_j|\forall r_j\in\mathcal{S}_3,\ dist(r_j, e_i)<D_{thres}\}$, $\forall r_i\in\mathcal{S}_3$, search rightwards in left view along $e_i\in\mathcal{E}_l$, calculate $\mathcal{L}_i=\{l_j|\forall l_j\in\mathcal{S}_2,\ dist(l_j, e_i)<D_{thres}\}$;\\
			\STATE Perform RoI alignment to $P_{l_i}$ and $P_{r_j}$, compute two sets of costs $\mathcal{C}_{li}=\{Prob_{ij}|\forall r_j\in\mathcal{R}_i, \mathcal{R}_i\neq\varnothing\}$ and $\mathcal{C}_{ri}=\{Prob_{ji}|\forall l_j\in\mathcal{L}_i, \mathcal{L}_i\neq\varnothing\}$;\\
			\STATE Find matches from either view. If $\mathcal{C}_{li}\neq\varnothing$ and $\exists r_j\in\mathcal{R}_i\ \mathrm{s.t.}\ Prob_{ij}\geq max(P_{thres},max(\mathcal{C}_{li}))$, let $\mathcal{M}_l= \mathcal{M}_l\cup\{(l_i, r_i)\}$. If $\mathcal{C}_{ri}\neq\varnothing$ and $\exists l_j\in\mathcal{L}_i\ \mathrm{s.t.}\ Prob_{ji}\geq max(P_{thres},max(\mathcal{C}_{ri}))$, let $\mathcal{M}_r= \mathcal{M}_r\cup\{(l_j, r_i)\}$; \\
			\RETURN $\mathcal{M}=\mathcal{M}_l\cap\mathcal{M}_r$;  
		\end{algorithmic}
	\end{algorithm}
	
	\begin{figure}[htbp]
		\centering
		\null\hfill
		\subfloat[]{
			\label{fig:5a}
			\includegraphics[width=0.45\linewidth]{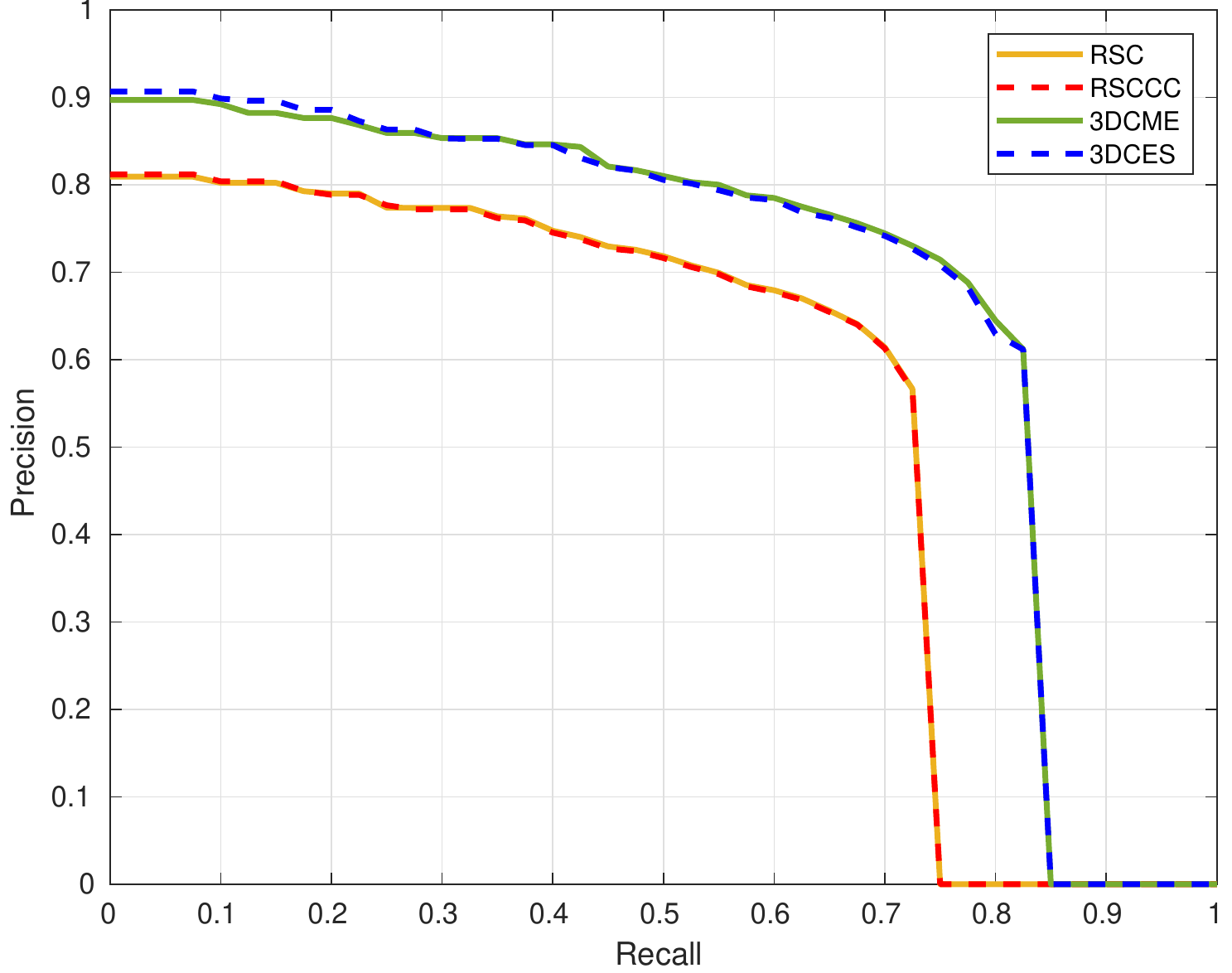}}
		\hfil
		\subfloat[]{
			\label{fig:5b}
			\includegraphics[width=0.45\linewidth]{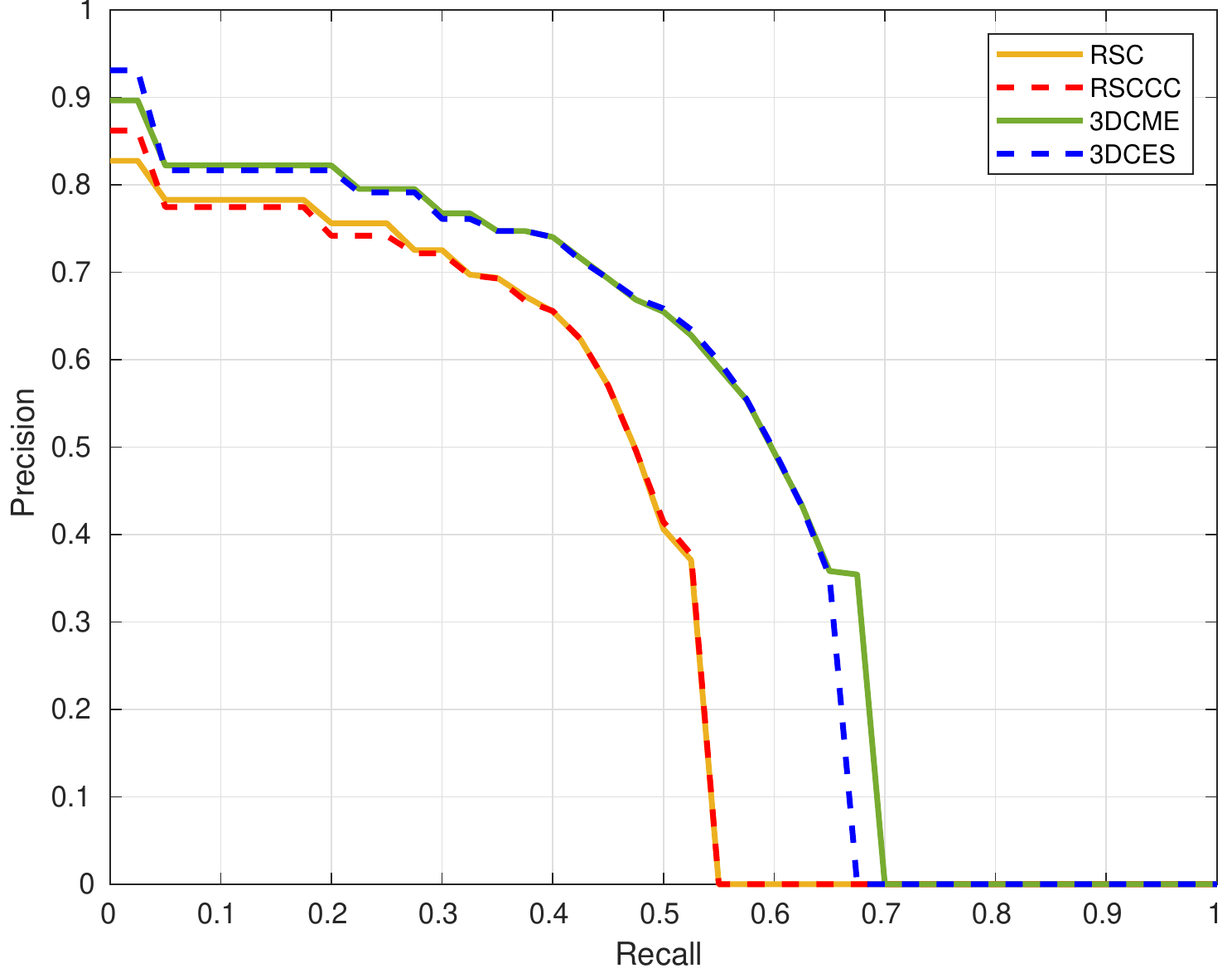}}
		\hfill\null
		\caption{P-R curves of 3D object detection results. \textbf{Moderate} level is presented. (a) Car category at IoU threshold 0.7. (b) Pedestrian category at IoU threshold 0.5.}
		\hfill\null
		\label{fig:5}
	\end{figure}
	
	\section{Experiments}\label{section:experiments}
	In this section, we present both validation and testing results on publicly available KITTI dataset. The dataset consists of 7,481 annotated (2D/3D bbox labeled in left view only) image pairs with corresponding point clouds and calibrations, 7518 unlabeled testing image pairs with corresponding point clouds and calibrations. In our experiments, 20\% of 7,481 randomly shuffled training samples are divided as the validation set, 2 categories of objects - `Car' and `Pedestrian' are examined for 3D/BEV detections. Also, we compare proposed SPFN with the state-of-the-arts including Pseudo-LiDAR \cite{wang2019pseudo}, Stereo R-CNN \cite{li2019stereo}, TLNet \cite{qin2019triangulation}, et. al.
	
	\subsection{Experiments Setup} \label{subsection:Setup}
	The 2D detector Faster R-CNN \cite{ren2015faster} (VGG-16 backbone) is trained on KITTI training set to provide shared weights and parameters for the sibling branches as shown in Fig.\ \ref{fig:3}, and Mask R-CNN \cite{he2017mask} (ResNet backbone + Feature Pyramid Network) is pre-trained on COCO dataset. We train FPNv2 with raw LiDAR data other than segmented point cloud from the matching module. In this way, each module pairs with FPNv2 as 3DCES-FPNv2, 3DCME-FPNv2, RSC-FPNv2, RSCCC-FPNv2. Before testing, several parameters need to be manually set: the percentage of bbox size $S_{enlarge}$ to be enlarged, distance threshold  $D_{thres}$ from the bbox center to the epipolar line, similarity threshold $P_{thres}$ for the template matching, the minimum number of inliers of stereo-frustums intersection $N_{thres}$ and $P_{3d\_thres}$ the 3D IoU threshold. We set $S_{enlarge}$ to 8\%, $N_{thres}$ to 5,  $D_{thres}$=30px, $P_{thres}$=0.4, and $P_{3d\_thres}$=0.5. In this section, we enforce enlargement to all bboxes unless specified by `NE' (w/o enlargement).
	
	\begin{table}[htbp]
		\centering
		\begin{tabular}{|c|c|c|c|}
			\hline
			2D Detector & $\mathrm{AP_{2D}}$ & $\mathrm{AP_{BEV}}$ & $\mathrm{AP_{3D}}$\\\hline\hline
			Faster R-CNN(NE) & 65.9 & 56.6 & 55.5\\
			Mask R-CNN(NE) & 85.8 & 74.2 & 73.5\\
			Faster R-CNN & 65.9 & 57.7 & 56.7\\
			Mask R-CNN & 85.8 & 83.0 & 82.2\\\hline
		\end{tabular}
		\caption{3D object detection on car category using two 2D detectors, APs(\%) at \textbf{Moderate} level with IoU=0.5. Enlarged bboxes of both views by 8\% on height and width. Only 3DCES-FPNv2 is tested.  \label{tab:1}}
	\end{table}
	
	It should be noted that the reported 2D detection results come from 2D detection stage other than the projection of the predicted 3D bounding box to the front view. Tab.\ {\ref{tab:1}} shows that when the 2D detection AP is low, enlarging bboxes slightly increases $\mathrm{AP_{BEV}}$ and $\mathrm{AP_{3D}}$ since the original bboxes are not precise enough to capture all keypoints. On the other hand, in terms of SFPN's serial pipeline structure, higher 2D detection AP may result in better 3D detection. We also observe that, if setting $P_{3d\_thres}$ to a larger threshold, the number of total matches drops greatly and vice versa, which behaves similar to 2D IoU during the detection phase.
	
	\subsection{Quantitative Evaluation on Validation Set} \label{subsection:Quantitative}
	
	The results of BEV and 3D object detection are presented in Tab.\ \ref{tab:6}, Tab.\ \ref{tab:2}, Tab.\ \ref{tab:3}, and Tab.\ \ref{tab:7}. Both RSC and RSCCC require RoI alignment during matching. RSC is designed to relax the one-to-one onto mapping constraint by merely finding matches to RoIs in the left view while RSCCC strictly follows this constraint. As shown in the detection results, RSCCC-FPNv2 achieves better performance than RSC-FPNv2 with IoU=0.7, which provides evidence for the effectiveness of this constraint. 3DCME depends on the proposed 3D IoU without RoI alignment or epipolar line search, which achieves better AP than 3DCES with epipolar line search. We will show in Section \ref{subsection:Runtime} that RSC runs faster than the other three matching methods, and 3DCME is the slowest but with the highest AP.
	
	\paragraph{Overview on comparisions} 3DCES-FPNv2 and 3DCME-FPNv2 achieve competitive performance among all state-of-the-art stereo-based methods because of highly precise LiDAR data. Although almost all listed stereo-based methods yield higher 2D detection AP (IoU=0.7) on either the validation set or the test set, 3DCES-FPNv2 and 3DCME-FPNv2 regress disproportional accurate 3D bboxes when projected to BEV. It is interesting that RSCCC-FPNv2 and RSC-FPNv2 outperform their learning-based versions (without LiDAR data) Stereo-RCNN ($\sim$10\%) and TLNet($\sim$35\%) by a large margin. 
	
	\begin{table}[htbp]
		\centering
		\begin{tabular}{|c|c|c|c|c|c|c|c|c|c|}
			\hline
			\multirow{2}{*}{Method} & \multirow{2}{*}{Type} & \multicolumn{4}{c|}{IoU=0.5} \\\cline{3-6}
			& & $\mathrm{AP_{2D}}$ & Easy & Mode & Hard \\\hline\hline
			RSC-FPNv2(\textbf{ours}) & Stereo+LiDAR & 85.8 & 71.9 & 68.4 & 61.5 \\
			RSCCC-FPNv2(\textbf{ours}) & Stereo+LiDAR & 85.8 & 71.9 & 68.2 & 61.4 \\
			3DCME-FPNv2(\textbf{ours}) & Stereo+LiDAR & 85.8 & 82.9 & 82.9 & 75.7 \\
			3DCES-FPNv2(\textbf{ours}) & Stereo+LiDAR & 85.8 & \textbf{83.1} & \textbf{83.0} & \textbf{75.7} \\
			PSMNET-AVOD \cite{wang2019pseudo} & Stereo & - & 89.0 & 77.5 & 68.7 \\
			Stereo R-CNN \cite{li2019stereo} & Stereo & - & 87.1 & 74.1 & 58.9 \\
			3DOP \cite{chen20173d} & Stereo & - & 55.0 & 41.3 & 34.6 \\
			TLNet \cite{qin2019triangulation} & Stereo & - & 62.5 & 46.0 & 41.9 \\\hline
		\end{tabular}
		\caption{BEV results on validation set. Car category is evaluated with IoU=0.5 and all APs in `\%'.  \textbf{Moderate} $\mathrm{AP_{2D}}$ on validation set is reported.\label{tab:6}}
	\end{table} 
	
	\begin{table}[htbp]
		\centering
		\begin{tabular}{|c|c|c|c|c|c|c|c|c|c}
			\hline
			\multirow{2}{*}{Method} & \multirow{2}{*}{Type} & \multicolumn{4}{c|}{IoU=0.7} \\
			\cline{3-6}
			& & $\mathrm{AP_{2D}}$ & Easy & Mode & Hard \\\hline\hline
			RSC-FPNv2(\textbf{ours}) & Stereo+LiDAR & 54.2 & 67.1 & 58.0 & 51.3 \\
			RSCCC-FPNv2(\textbf{ours}) & Stereo+LiDAR & 54.2 & 68.4 & 58.6 & 51.8 \\
			3DCME-FPNv2(\textbf{ours}) & Stereo+LiDAR & 54.2 & \textbf{79.8} & \textbf{72.9} & \textbf{65.7}\\
			3DCES-FPNv2(\textbf{ours}) & Stereo+LiDAR & 54.2 & 78.4 & 72.2 & 65.1 \\
			PSMNET-AVOD \cite{wang2019pseudo} & Stereo & - & 74.9 & 56.8 & 49.0 \\
			Stereo R-CNN \cite{li2019stereo} & Stereo & \textbf{88.3} & 68.5 & 48.3 & 41.5 \\
			3DOP \cite{chen20173d} & Stereo & - & 55.0 & 9.5 & 7.6 \\
			TLNet \cite{qin2019triangulation} & Stereo & - & 29.22 & 21.9 & 18.8 \\\hline
		\end{tabular}
		\caption{BEV results on validation set. Car category is evaluated with IoU=0.7 and all APs in `\%'. \textbf{Moderate} $\mathrm{AP_{2D}}$ on validation set is reported.\label{tab:2}}
	\end{table} 
	
	\begin{table}[htbp]
		\centering
		\begin{tabular}{|c|c|c|c|c|c|c|c|c|c|}
			\hline
			\multirow{2}{*}{Method} & \multirow{2}{*}{Type} & \multicolumn{4}{c|}{IoU=0.5} \\
			\cline{3-6}
			& & $\mathrm{AP_{2D}}$ & Easy & Mode & Hard \\\hline\hline
			RSC-FPNv2(\textbf{ours}) & Stereo+LiDAR & 85.8 & 70.7 & 67.1 & 54.0 \\
			RSCCC-FPNv2(\textbf{ours}) & Stereo+LiDAR & 85.8 & 70.4 & 66.9 & 53.9 \\
			3DCME-FPNv2(\textbf{ours}) & Stereo+LiDAR & 85.8 & 82.5 & \textbf{82.3} & \textbf{75.0}\\
			3DCES-FPNv2(\textbf{ours}) & Stereo+LiDAR & 85.8 & 82.4 & 82.2 & 74.9 \\
			PSMNET-AVOD \cite{wang2019pseudo} & Stereo & - & \textbf{88.5} & 76.4 & 61.2 \\
			Stereo R-CNN \cite{li2019stereo} & Stereo & - & 85.8 & 66.3 & 57.2 \\
			3DOP \cite{chen20173d} & Stereo & - & 46.0 & 34.6 & 30.1 \\
			TLNet \cite{qin2019triangulation} & Stereo & - & 59.5 & 43.7 & 38.0 \\\hline
		\end{tabular}
		\caption{3D detection results on validation set. Car category is evaluated with IoU=0.5. 3DCME-FPNv2 achieves the best performance on validation set out of stereo-based methods.\label{tab:3}}
	\end{table}
	
	\begin{table}[htbp]
		\centering
		\begin{tabular}{|c|c|c|c|c|c|c|c|c|c|}
			\hline
			\multirow{2}{*}{Method} & \multirow{2}{*}{Type} & \multicolumn{4}{c|}{IoU=0.7} \\
			\cline{3-6}
			& & $\mathrm{AP_{2D}}$ & Easy & Mode & Hard \\\hline\hline
			RSC-FPNv2(\textbf{ours}) & Stereo+LiDAR & 54.2 & 58.5 & 53.7 & 47.4 \\
			RSCCC-FPNv2(\textbf{ours}) & Stereo+LiDAR & 54.2 & 58.7 & 53.9 & 47.5 \\
			3DCME-FPNv2(\textbf{ours}) & Stereo+LiDAR & 54.2 & \textbf{73.0} & \textbf{67.3} & \textbf{61.2}\\
			3DCES-FPNv2(\textbf{ours}) & Stereo+LiDAR & 54.2 & 72.5 & 66.7 & 60.9 \\
			PSMNET-AVOD \cite{wang2019pseudo} & Stereo & - & 61.9 & 45.3 & 39.0 \\
			Stereo R-CNN \cite{li2019stereo} & Stereo & \textbf{88.3} & 54.1 & 36.7 & 31.1 \\
			3DOP \cite{chen20173d} & Stereo & - & 6.6 & 5.1 & 4.1 \\
			TLNet \cite{qin2019triangulation} & Stereo & - & 18.2 & 14.3 & 13.7 \\\hline
		\end{tabular}
		\caption{3D detection results on validation set. Car category is evaluated with IoU=0.7. 3DCME-FPNv2 achieves the best performance on validation set out of stereo-based methods.\label{tab:7}}
	\end{table}
	
	\paragraph{Comparision with Pseudo-LiDAR} Tab.\ \ref{tab:5} shows the comparison with a recently published Pseudo-LiDAR \cite{wang2019pseudo} method. Compared with FPNv2 on validation set and the car category, 3DCES-FPNv2 and 3DCME-FPNv2 achieve competitive results although they suffer from lower recall (see Fig.\ \ref{fig:5a}) considering mismatches in both views. In order to explore potentials of proposed method, we specifically select PSMNET-AVOD that achieves the best performance among the methods proposed in \cite{wang2019pseudo}. As mentioned in Section \ref{section:intro}, dense disparity map generates a coarse estimation of the scene which is sensitive to the focal length and baseline of the stereo cameras. According to Pseudo-LiDAR \cite{wang2019pseudo}, its point cloud detector AVOD and FPN are trained by fine-grained, sparse LiDAR points, therefore its performance may not compete with the state-of-the-art methods based on real-LiDAR data.
	
	\begin{table}[htbp]
		\centering
		\begin{tabular}{|c|c|c|c|c|}
			\hline
			Method & $\mathrm{AP_{2D}}$ & Easy & Mode & Hard \\\hline\hline
			3DCME-FPNv2(\textbf{ours}) & 57.4 & 57.6 & 47.1 & 40.7 \\
			3DCES-FPNv2(\textbf{ours}) & 57.4 & \textbf{57.8} & 47.3 & 40.8 \\
			FPNv2(\textbf{our results}) & 57.4 & 55.6 & \textbf{49.1} & \textbf{42.6} \\\hline
			PSMNET-FPN \cite{wang2019pseudo} & - & 33.8 & 27.4 & 24.0 \\
			FPN \cite{wang2019pseudo} & - & 64.7 & 56.5 & 49.9 \\\hline
		\end{tabular}
		\caption{3D detection results on validation set. Pedestrian category is evaluated.\label{tab:5}}
	\end{table}
	
	\begin{table}[htbp]
		\centering
		\begin{tabular}{|c|c|c|c|c|}
			\hline
			Method & $\mathrm{AP_{2D}}$ & Easy & Mode & Hard \\\hline\hline
			3DCME-FPNv2(\textbf{ours}) & 57.4 & \textbf{62.6} & \textbf{53.7} & \textbf{46.6} \\
			3DCES-FPNv2(\textbf{ours}) & 57.4 & 60.8 & 52.8 & 45.9 \\
			FPNv2(\textbf{our results}) & 57.4 & 58.5 & 51.5 & 44.8 \\\hline
			PSMNET-FPN \cite{wang2019pseudo} & - & 41.3 & 34.9 & 30.1 \\
			FPN \cite{wang2019pseudo} & - & 69.7 & 60.6 & 53.4 \\\hline
		\end{tabular}
		\caption{BEV detection results on validation set. Pedestrian category is evaluated. Results of FPN in the table are presented in \cite{wang2019frustum}.\label{tab:4}}
	\end{table}
	
	Tab.\ \ref{tab:4} shows the BEV results compared to FPNv2 on validation set, along with another similar comparison with \cite{wang2019pseudo}. 3DCME-FPNv2 achieves better performance than FPNv2 by 4.1\% at Easy level, 2.2\% at Moderate level, and 1.8\% at Hard level. The improvements are mainly derived from enlarged bboxes since they enrich sparse segmentation captured by stereo frustums. These enlarged bboxes are especially efficient for pedestrians whose poses vary. For 3D object detection, 3DCES-FPNv2 achieves higher AP at Easy level, but is less precise than FPNv2 at Moderate and Hard levels. To conclude, when comparing margins to corresponding FPN detection results in Tab.\ \ref{tab:4} and Tab.\ \ref{tab:5}, our methods outperform pseudo-LiDAR based PSMNET-FPN by significant margins.
	
	\begin{figure}[htbp]
		\centering
		\includegraphics[width=\linewidth]{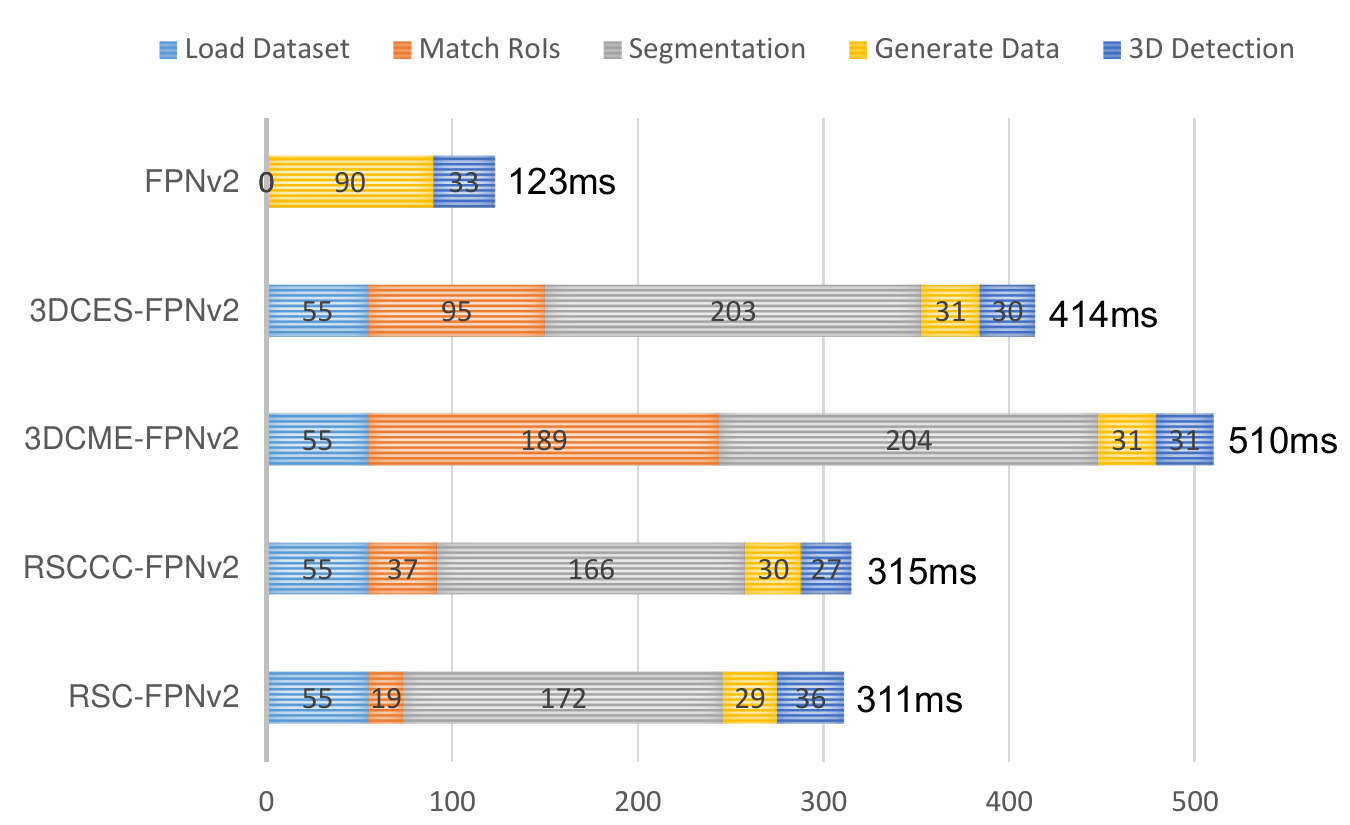}
		\DeclareGraphicsExtensions.
		\caption{Runtime comparison between matching modules and original FPNv2, best viewed in colour mode. Processing time (millionsecond) is calculated by averaging runtime of validating 1,496 samples.}
		\label{fig:4}
	\end{figure}
	
	\begin{figure}[htbp]
		\centering
		\subfloat[3D(left) and BEV(right) detections. More persons than cars.]{
			\label{fig:6a}
			\includegraphics[width=0.95\linewidth]{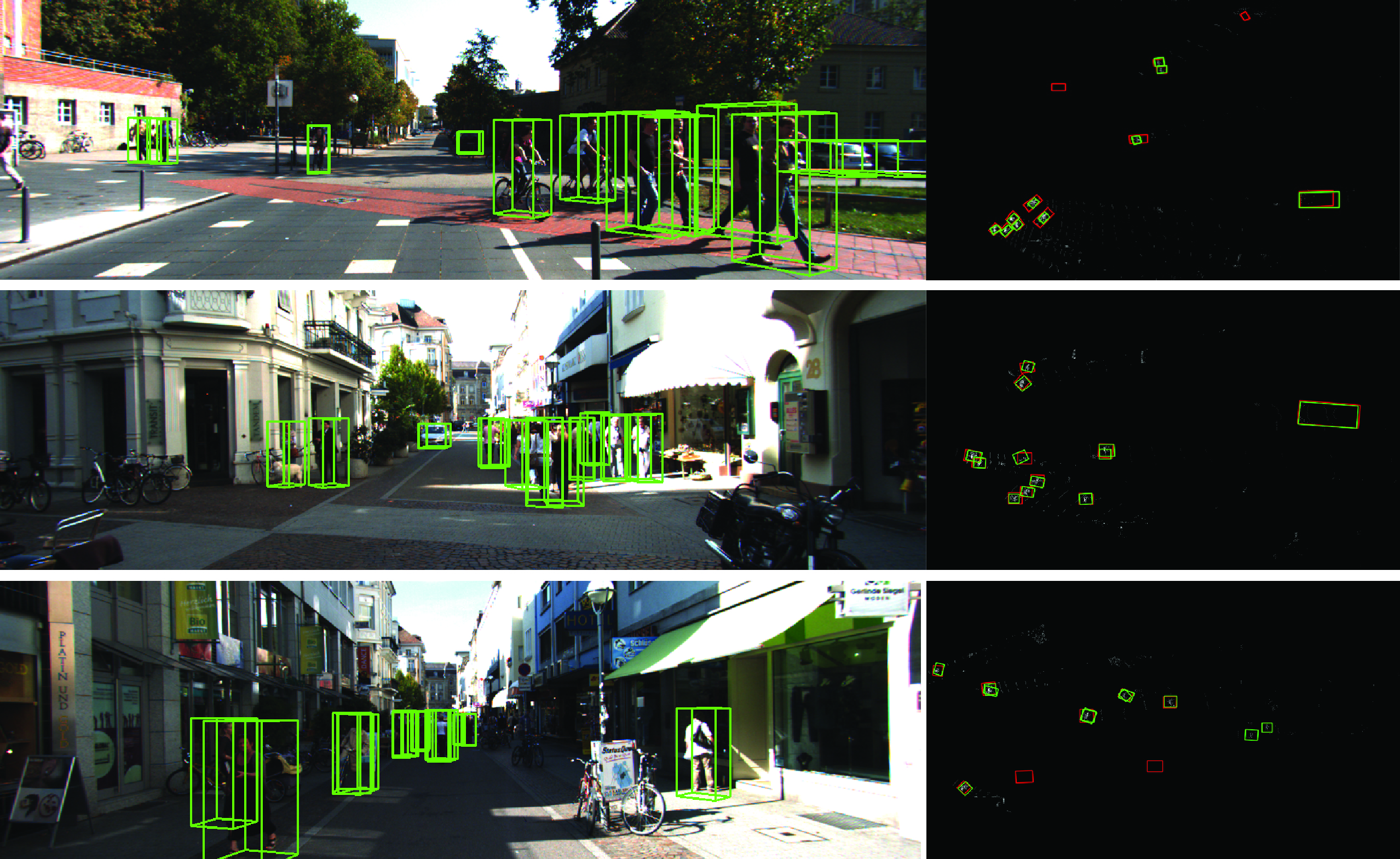}}
		\newline
		\subfloat[3D(left) and BEV(right) detections. \textbf{Car} category only.]{
			\label{fig:6b}
			\includegraphics[width=0.95\linewidth]{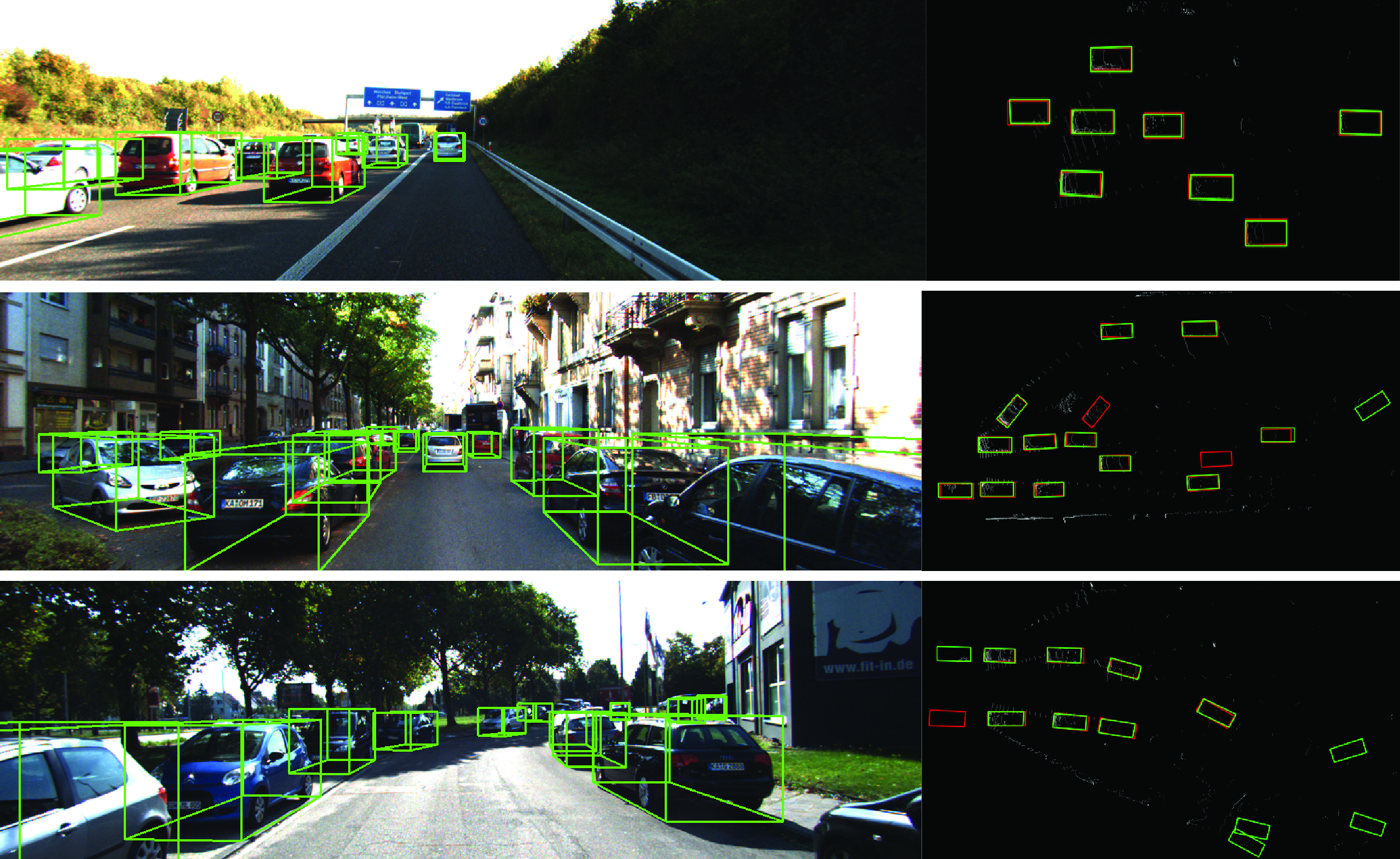}}
		\caption{Qualitative evaluation on validation set using Mask R-CNN detector. Red bboxes in BEV indicate ground truths, and green bboxes indicate detections. Cyclist category is not evaluated in our experiments. However, Mask R-CNN \cite{he2017mask} misclassifies cyclists as pedestrians, therefore significantly increases false positive rate in detecting pedestrians.}
		\label{fig:6}
	\end{figure}
	
	\subsection{Quantitative Evaluation on Test Set} \label{subsection:Test}
	
	In this subsection, we present the performance of 3DCES-FPNv2 on KITTI leaderboard \cite{KITTI} among all stereo-based submissions. We take Mask R-CNN \cite{he2017mask} as the 2D detector as it has a higher detection AP over Faster R-CNN (see Section \ref{subsection:Setup}). FPNv2 is trained on our training set other than all 7,481 samples. Tab.\ \ref{tab:9} and Tab.\ \ref{tab:11} present 3D object detection results, Tab.\ \ref{tab:10} and Tab.\ \ref{tab:12} present BEV detection results.
	
	It can be observed that from Tab.\ \ref{tab:11} and Tab.\ \ref{tab:12}, proposed method outperforms stereo-based method in detecting pedestrians. However, as there are more samples in the cyclist category than that of the validation set, while Mask R-CNN mistakes the cyclist category as the pedestrian category, we believe this is one of the main factors that lower the detection APs. In terms of the performance in 3D object detection, 3DCES-FPNv2 achieves competitive performance among stereo-based methods. It should be noted that untrained Mask R-CNN provides much lower 2D detection AP at 46.68\% which is 38.47\% less than 85.15\% of PL++(SDN+GDC) \cite{you2019pseudo}. Nonetheless, the proposed method maintains disproportionate performance in 3D object detection, which proves the effectiveness of the proposed pipeline.
	\begin{table}[htbp]
		\centering
		\begin{tabular}{|c|c|c|c|c|c|}
			\hline
			Method & Type & $\mathrm{AP_{2D}}$ & Easy & Mode & Hard \\\hline\hline
			3DCES-FPNv2(\textbf{ours}) & Stereo+LiDAR & \textbf{46.68} & \textbf{58.88} & \textbf{51.92} & \textbf{44.59} \\
			Pseudo-LiDAR \cite{wang2019pseudo} & Stereo & 67.79 & 54.53 & 34.05 & 28.25 \\
			Pseudo-LiDAR++ \cite{you2019pseudo} & Stereo & 82.90 & 61.11 & 42.43 & 36.99 \\
			PL++(SDN+GDC) \cite{you2019pseudo} & Stereo+LiDAR & \textbf{85.15} & \textbf{68.38} & \textbf{54.88} & \textbf{49.16}\\ 
			ZoomNet \cite{xu2020zoomnet} & Stereo & 83.92 & 55.98 & 38.64 & 30.97\\
			Stereo R-CNN \cite{li2019stereo} & Stereo & 85.98 & 47.58 & 30.23 & 23.72 \\
			StereoFENet \cite{bao2019monofenet} & Stereo & 85.70 & 29.14 & 18.41 & 14.20 \\
			OC Stereo \cite{pon2020object} & Stereo & 74.60 & 55.15 & 37.60 & 30.25\\
			RT3DStereo \cite{Knigshof2019Realtime3D} & Stereo & 45.81 & 29.90 & 23.28 & 18.96\\
			TLNet \cite{qin2019triangulation} & Stereo & 63.53 & 7.64 & 4.37 & 3.74\\
			\hline
		\end{tabular}
		\caption{3D detection APs(\%) on test set\cite{KITTI}. Car category is evaluated. \textbf{Moderate} 2D detection APs are reported. With much lower 2D detection AP, proposed method has the potential to outperform PL++(SDN+GDC).\label{tab:9}}
	\end{table}
	
	\begin{table}[htbp]
		\centering
		\begin{tabular}{|c|c|c|c|c|c|}
			\hline
			Method & Type & $\mathrm{AP_{2D}}$ & Easy & Mode & Hard \\\hline\hline
			3DCES-FPNv2(\textbf{ours}) & Stereo+LiDAR & \textbf{51.83} & \textbf{37.16} & \textbf{29.77} & \textbf{26.61} \\
			OC Stereo \cite{pon2020object} & Stereo & 30.79 & 29.79 & 20.80 & 18.62 \\
			RT3Dstereo \cite{Knigshof2019Realtime3D} & Stereo & 29.30 & 4.72 & 3.65 & 3.00\\
			\hline
		\end{tabular}
		\caption{3D detection APs(\%) on test set \cite{KITTI}. Pedestrian category is evaluated.\label{tab:11}}
	\end{table}
	
	\begin{table}[htbp]
		\centering
		\begin{tabular}{|c|c|c|c|c|c|}
			\hline
			Method & Type & $\mathrm{AP_{2D}}$ & Easy & Mode & Hard \\\hline\hline
			3DCES-FPNv2(\textbf{ours}) & Stereo+LiDAR & \textbf{46.68} & \textbf{74.20} & \textbf{65.74} & \textbf{58.35} \\
			Pseudo-LiDAR \cite{wang2019pseudo} & Stereo & 67.79 & 67.30 & 45.00 & 38.40 \\
			Pseudo-LiDAR++ \cite{you2019pseudo} & Stereo & 82.90 & 78.31 & 58.01 & 51.25 \\
			PL++(SDN+GDC) \cite{you2019pseudo} & Stereo+LiDAR & \textbf{85.15} & \textbf{84.61} & \textbf{73.80} & \textbf{65.59}\\ 
			ZoomNet \cite{xu2020zoomnet} & Stereo & 83.92 & 72.94 & 54.91 & 44.14 \\
			Stereo R-CNN \cite{li2019stereo} & Stereo & 85.98 & 61.92 & 41.31 & 33.42\\
			StereoFENet \cite{bao2019monofenet} & Stereo & 85.70 & 49.29 & 32.96 & 25.90 \\
			OC Stereo \cite{pon2020object} & Stereo & 74.60 & 68.89 & 51.47 & 42.97 \\
			RT3Dstereo \cite{Knigshof2019Realtime3D} & Stereo & 45.81 & 58.81 & 46.82 & 38.38\\
			TLNet \cite{qin2019triangulation} & Stereo & 63.53 & 13.71 & 7.69 & 6.73\\
			\hline
		\end{tabular}
		\caption{BEV detection APs(\%) on test set \cite{KITTI}. Car category is evaluated.\label{tab:10}}
	\end{table}
	
	\begin{table}[htbp]
		\centering
		\begin{tabular}{|c|c|c|c|c|c|}
			\hline
			Method & Type & $\mathrm{AP_{2D}}$ & Easy & Mode & Hard \\\hline\hline
			3DCES-FPNv2(\textbf{ours}) & Stereo+LiDAR & \textbf{51.83} & \textbf{31.61} & \textbf{24.84} & \textbf{21.96} \\
			OC Stereo \cite{pon2020object} & Stereo & 30.79 & 24.48 & 17.58 & 15.60 \\
			RT3Dstereo \cite{Knigshof2019Realtime3D} & Stereo & 29.30 & 3.28 & 2.45 & 2.35\\
			\hline
		\end{tabular}
		\caption{BEV detection APs(\%) on test set \cite{KITTI}. Pedestrian category is evaluated.\label{tab:12}}
	\end{table}
	
	\subsection{Runtime} \label{subsection:Runtime}
	
	With a fixed number of LiDAR points (e.g., 1024) fed into FPNv2, 3D detection phases have almost the same computation costs (see Fig.\ \ref{fig:4}). Therefore, it is necessary to show runtime efficiency of the stereo frustums pipeline. As depicted in Fig.\ \ref{fig:4}, all shown phases are tested with a 2.5Ghz CPU except for the 3D detection phase which is tested with a P100 GPU. RSC-FPNv2 is the fastest among the four methods, RSCCC-FPNv2 doubles the matching time of RSC-FPNv2 but only slightly increases the overall preparation time, 3DCME significantly increases RoIs matching time while 3DCES reduces the time almost by half. Thus, 3DCES is efficient in matching as well as maintaining detection AP as 3DCME does. Most runtime of data-preparation phase goes to point cloud processing, which remains to be optimized by high-performance computing techniques.

	\subsection{Qualitative Results} \label{subsection:quality}
	
	The results of 3D detection and BEV detection using SPFN on the validation set are visualized in Fig.\ \ref{fig:6}. The BEV detections are depicted on a sparse point cloud generated by the 3DCES matching module. 
	Fig.\ \ref{fig:6a} shows that two pedestrians are missed due to unsatisfying illumination condition that invalids the 2D detector, another two pedestrians are missed due to large occlusion and too small in dimensions. Despite the missed detections, SFPN is capable of regressing highly precise bboxes that enclose irregular objects as pedestrians. There is a very close object in Fig.\ \ref{fig:6b} which is not detected due to too few clues of objectness in that region. Also, largely occluded objects in Fig.\ \ref{fig:6b} can hardly be detected. Nevertheless, RoIs locate both very near and near (around 40\textit{m}, which almost reaches the valid-range limit of LiDAR sensor) objects precisely. Some faraway objects located at 50-70\textit{m} can be detected by SFPN but with less precision. 
	To further study segmented point cloud of the faraway objects, the number of points in each segmented point cloud is usually less than 100, we believe this number is around the borderline threshold $N_{thres}$.
	
	\section{Conclusion}
	
	In this paper, we have proposed four matching modules to bridge the gap between 2D object detection on stereopsis and real LiDAR data via dense epipolar geometry constraints: the one-to-one onto mapping and minimum intersection. To accommodate the proposed matching modules, we have proposed a stereo frustum pipeline for 3D object detection where the 2D detection results are fed to the matching module to generate matches to segment the point cloud of the scene, and the 3D segmentation proposals are then fed to a refinement network for more precise objectness segmentation, followed by a multi-modal regression network. 
	
	By integrating with the F-PointNets, our stereo frustum pipeline can achieve 2-3 frames per second without coding optimization. Although this frame rate is not yet applicable for real-time applications, its speed can be further increased if we adopt techniques like GPU acceleration. More efficient matching algorithms and 2D detection models are also expected. The proposed pipeline outperforms the state-of-the-art stereo-based approaches with a lower 2D detection average precision, it has the potential to outperform the state-of-the-art LiDAR and stereo fusion approaches if better 2D detection models are adopted.
	
	We are currently working to design end-to-end matching modules for 3DCES and 3DCME to achieve more effective representation to encode the sparse point cloud. Some future work includes leveraging the reliability of both views for better performance in the detection accuracy and recall, optimizing runtime not only at the coding level, but also seeking into the possibilities of implementing distributed parallel computing techniques.
	
	
	
	\bibliographystyle{spmpsci}
	\bibliography{mybibfile}

\begin{thebibliography}{10}
\providecommand{\url}[1]{{#1}}
\providecommand{\urlprefix}{URL }
\expandafter\ifx\csname urlstyle\endcsname\relax
  \providecommand{\doi}[1]{DOI~\discretionary{}{}{}#1}\else
  \providecommand{\doi}{DOI~\discretionary{}{}{}\begingroup
  \urlstyle{rm}\Url}\fi

\bibitem{bao2019monofenet}
Bao, W., Xu, B., Chen, Z.: Monofenet: Monocular 3d object detection with
  feature enhancement networks.
\newblock Transactions on Image Processing  (2019)

\bibitem{chen20173d}
Chen, X., Kundu, K., Zhu, Y., Ma, H., Fidler, S., Urtasun, R.: 3d object
  proposals using stereo imagery for accurate object class detection.
\newblock Transactions on Pattern Analysis and Machine Intelligence
  \textbf{40}(5), 1259--1272 (2017)

\bibitem{du2018general}
Du, X., Ang, M.H., Karaman, S., Rus, D.: A general pipeline for 3d detection of
  vehicles.
\newblock In: IEEE International Conference on Robotics and Automation, pp.
  3194--3200 (2018)

\bibitem{geiger2013vision}
Geiger, A., Lenz, P., Stiller, C., Urtasun, R.: Vision meets robotics: The
  kitti dataset.
\newblock The International Journal of Robotics Research \textbf{32}(11),
  1231--1237 (2013)

\bibitem{KITTI}
Geiger, A., Lenz, P., Urtasun, R.: Official {KITTI} benchmark.
\newblock \urlprefix\url{http://www.cvlibs.net/datasets/kitti/}.
\newblock Accessed: 2019-11-19

\bibitem{he2017mask}
He, K., Gkioxari, G., Doll{\'a}r, P., Girshick, R.: Mask r-cnn.
\newblock In: IEEE International Conference on Computer Vision, pp. 2961--2969
  (2017)

\bibitem{Knigshof2019Realtime3D}
K{\"o}nigshof, H., Salscheider, N.O., Stiller, C.: Realtime 3 d object
  detection for automated driving using stereo vision and semantic information.
\newblock In: IEEE International Conference on Intelligent Transportation
  Systems (2019)

\bibitem{ku2018joint}
Ku, J., Mozifian, M., Lee, J., Harakeh, A., Waslander, S.L.: Joint 3d proposal
  generation and object detection from view aggregation.
\newblock In: IEEE/RSJ International Conference on Intelligent Robots and
  Systems, pp. 1--8 (2018)

\bibitem{li20202}
Li, K., Ma, W., Sajid, U., Wu, Y., Wang, G.: 2 object detection with
  convolutional neural networks.
\newblock Deep Learning in Computer Vision: Principles and Applications
  \textbf{30}(31), 41 (2020)

\bibitem{li2019stereo}
Li, P., Chen, X., Shen, S.: Stereo r-cnn based 3d object detection for
  autonomous driving.
\newblock In: IEEE Conference on Computer Vision and Pattern Recognition, pp.
  7644--7652 (2019)

\bibitem{liang2018deep}
Liang, M., Yang, B., Wang, S., Urtasun, R.: Deep continuous fusion for
  multi-sensor 3d object detection.
\newblock In: European Conference on Computer Vision, pp. 641--656 (2018)

\bibitem{lin2017focal}
Lin, T.Y., Goyal, P., Girshick, R., He, K., Doll{\'a}r, P.: Focal loss for
  dense object detection.
\newblock In: IEEE International Conference on Computer Vision, pp. 2980--2988
  (2017)

\bibitem{luo2018fast}
Luo, W., Yang, B., Urtasun, R.: Fast and furious: Real time end-to-end 3d
  detection, tracking and motion forecasting with a single convolutional net.
\newblock In: IEEE conference on Computer Vision and Pattern Recognition, pp.
  3569--3577 (2018)

\bibitem{ma2020mdfn}
Ma, W., Wu, Y., Cen, F., Wang, G.: Mdfn: Multi-scale deep feature learning
  network for object detection.
\newblock Pattern Recognition \textbf{100}, 107149 (2020)

\bibitem{pon2020object}
Pon, A.D., Ku, J., Li, C., Waslander, S.L.: Object-centric stereo matching for
  3d object detection.
\newblock In: IEEE International Conference on Robotics and Automation, pp.
  8383--8389 (2020)

\bibitem{qi2018frustum}
Qi, C.R., Liu, W., Wu, C., Su, H., Guibas, L.J.: Frustum pointnets for 3d
  object detection from rgb-d data.
\newblock In: IEEE Conference on Computer Vision and Pattern Recognition, pp.
  918--927 (2018)

\bibitem{qi2017pointnet}
Qi, C.R., Su, H., Mo, K., Guibas, L.J.: Pointnet: Deep learning on point sets
  for 3d classification and segmentation.
\newblock In: IEEE Conference on Computer Vision and Pattern Recognition, pp.
  652--660 (2017)

\bibitem{qi2017pointnet++}
Qi, C.R., Yi, L., Su, H., Guibas, L.J.: Pointnet++: Deep hierarchical feature
  learning on point sets in a metric space.
\newblock In: Advances in Neural Information Processing Systems, pp. 5099--5108
  (2017)

\bibitem{qin2019triangulation}
Qin, Z., Wang, J., Lu, Y.: Triangulation learning network: from monocular to
  stereo 3d object detection.
\newblock In: IEEE Conference on Computer Vision and Pattern Recognition, pp.
  7607--7615 (2019)

\bibitem{ren2015faster}
Ren, S., He, K., Girshick, R., Sun, J.: Faster r-cnn: Towards real-time object
  detection with region proposal networks.
\newblock In: Advances in Neural Information Processing Systems, pp. 91--99
  (2015)

\bibitem{shi2019pointrcnn}
Shi, S., Wang, X., Li, H.: Pointrcnn: 3d object proposal generation and
  detection from point cloud.
\newblock In: IEEE Conference on Computer Vision and Pattern Recognition, pp.
  770--779 (2019)

\bibitem{shi2020points}
Shi, S., Wang, Z., Shi, J., Wang, X., Li, H.: From points to parts: 3d object
  detection from point cloud with part-aware and part-aggregation network.
\newblock Transactions on Pattern Analysis and Machine Intelligence  (2020)

\bibitem{shin2019roarnet}
Shin, K., Kwon, Y.P., Tomizuka, M.: Roarnet: A robust 3d object detection based
  on region approximation refinement.
\newblock In: IEEE Intelligent Vehicles Symposium (IV), pp. 2510--2515 (2019)

\bibitem{tian2018robust}
Tian, L., Li, M., Hao, Y., Liu, J., Zhang, G., Chen, Y.Q.: Robust 3-d human
  detection in complex environments with a depth camera.
\newblock Transactions on Multimedia \textbf{20}(9), 2249--2261 (2018)

\bibitem{wang2019voxel}
Wang, B., An, J., Cao, J.: Voxel-fpn: multi-scale voxel feature aggregation in
  3d object detection from point clouds.
\newblock Sensors \textbf{20}(3), 704 (2020)

\bibitem{wang2019pseudo}
Wang, Y., Chao, W.L., Garg, D., Hariharan, B., Campbell, M., Weinberger, K.Q.:
  Pseudo-lidar from visual depth estimation: Bridging the gap in 3d object
  detection for autonomous driving.
\newblock In: IEEE Conference on Computer Vision and Pattern Recognition, pp.
  8445--8453 (2019)

\bibitem{wang2019frustum}
Wang, Z., Jia, K.: Frustum convnet: Sliding frustums to aggregate local
  point-wise features for amodal.
\newblock In: IEEE/RSJ International Conference on Intelligent Robots and
  Systems, pp. 1742--1749 (2019)

\bibitem{xu2018pointfusion}
Xu, D., Anguelov, D., Jain, A.: Pointfusion: Deep sensor fusion for 3d bounding
  box estimation.
\newblock In: IEEE Conference on Computer Vision and Pattern Recognition, pp.
  244--253 (2018)

\bibitem{xu2020zoomnet}
Xu, Z., Zhang, W., Ye, X., Tan, X., Yang, W., Wen, S., Ding, E., Meng, A.,
  Huang, L.: Zoomnet: Part-aware adaptive zooming neural network for 3d object
  detection.
\newblock In: AAAI, pp. 12557--12564 (2020)

\bibitem{yang2018pixor}
Yang, B., Luo, W., Urtasun, R.: Pixor: Real-time 3d object detection from point
  clouds.
\newblock In: IEEE Conference on Computer Vision and Pattern Recognition, pp.
  7652--7660 (2018)

\bibitem{yang2018ipod}
Yang, Z., Sun, Y., Liu, S., Shen, X., Jia, J.: Ipod: Intensive point-based
  object detector for point cloud.
\newblock arXiv preprint arXiv:1812.05276  (2018)

\bibitem{yang2019std}
Yang, Z., Sun, Y., Liu, S., Shen, X., Jia, J.: Std: Sparse-to-dense 3d object
  detector for point cloud.
\newblock In: IEEE International Conference on Computer Vision, pp. 1951--1960
  (2019)

\bibitem{you2019pseudo}
You, Y., Wang, Y., Chao, W.L., Garg, D., Pleiss, G., Hariharan, B., Campbell,
  M., Weinberger, K.Q.: Pseudo-lidar++: Accurate depth for 3d object detection
  in autonomous driving.
\newblock In: International Conference on Learning Representations (2019)

\bibitem{zhang2017joint}
Zhang, G., Liu, J., Li, H., Chen, Y.Q., Davis, L.S.: Joint human detection and
  head pose estimation via multistream networks for rgb-d videos.
\newblock Signal Processing Letters \textbf{24}(11), 1666--1670 (2017)

\bibitem{Zhou2018}
Zhou, Y., Tuzel, O.: Voxelnet: End-to-end learning for point cloud based 3d
  object detection.
\newblock In: IEEE Conference on Computer Vision and Pattern Recognition, pp.
  4490--4499 (2018)

\end{thebibliography}
	
\end{document}